\documentclass[10pt,twocolumn,letterpaper]{article}

\usepackage{cvpr} 
\usepackage{times}
\usepackage{epsfig}
\usepackage{graphicx}
\usepackage{amsmath}
\usepackage{amssymb}

\usepackage[acronym,nopostdot]{glossaries}
\usepackage{booktabs} 
\usepackage{multirow} 

\usepackage[pagebackref]{hyperref}

\makeglossaries

\newacronym{cnn}{CNN}{Convolutional Neural Network}
\newacronym{pad}{PAD}{Presentation Attack Detection}
\newacronym{pa}{PA}{Presentation Attack}
\newacronym{kyc}{KYC}{Know Your Customer}
\newacronym{pai}{PAI}{Presentation Attack Instrument}
\newacronym{pais}{PAIS}{Presentation Attack Instrument Species}
\newacronym{gan}{GANs}{Generative Adversarial Networks}
\newacronym{dm}{DM}{Diffusion Models}
\newacronym{gdpr}{GDPR}{General Data Protection Regulation}
\newacronym{vit}{ViT}{Vision Transformers}
\newacronym{hdr}{HDR}{High Dynamic Range}
\newacronym{ppi}{PPI}{Pixels Per Inch}
\newacronym{dpi}{DPI}{Dots Per Inch}
\newacronym{gimp}{GIMP}{GNU Image Manipulator Program}
\newacronym{bpcer}{BPCER}{Bona-fide Presentation Classification Error Rate}
\newacronym{apcer}{APCER}{Attack Presentation Classification Error Rate}
\newacronym{acer}{ACER}{Average Classification Error Rate}
\newacronym{far}{FAR}{False Acceptance Rate}
\newacronym{frr}{FRR}{False Rejection Rate}
\newacronym{eer}{EER}{Equal Error Rate}
\newacronym{det}{DET}{Detection Error Trade-off}
\newacronym{bce}{BCE}{Binary Cross Entropy}
\newacronym{ai}{AI}{Artificial Intelligence}
\newacronym{pvc}{PVC}{Poly-Vinyl Chloride}

\begin{document}

\title{FakeIDet: Exploring Patches for Privacy-Preserving Fake ID Detection}
 
\author{
\begin{tabular}{c}
Javier Muñoz-Haro \quad Ruben Tolosana \quad Ruben Vera-Rodriguez \\
Aythami Morales \quad Julian Fierrez \\
\end{tabular} \\
{\normalsize Biometrics and Data Pattern Analytics Lab, Universidad Autonoma de Madrid, Madrid, Spain} \\
{\tt\small \{javier.munnoz, ruben.tolosana, ruben.vera, aythami.morales, julian.fierrez\}@uam.es}
}

\maketitle
\thispagestyle{empty}

\begin{abstract}
Verifying the authenticity of identity documents (IDs) has become a critical challenge for real-life applications such as digital banking, crypto-exchanges, renting, etc. This study focuses on the topic of fake ID detection, covering several limitations in the field. In particular, there are no publicly available data from real IDs for proper research in this area, and most published studies rely on proprietary internal databases that are not available for privacy reasons. In order to advance this critical challenge of real data scarcity that makes it so difficult to advance the technology of machine learning-based fake ID detection, we introduce a new patch-based methodology that trades off privacy and performance, and propose a novel patch-wise approach for privacy-aware fake ID detection: FakeIDet. In our experiments, we explore: \textit{i)} two levels of anonymization for an ID (i.e., fully- and pseudo-anonymized), and \textit{ii)} different patch size configurations, varying the amount of sensitive data visible in the patch image. State-of-the-art methods, such as vision transformers and foundation models, are considered as backbones. Our results show that, on an unseen database (DLC-2021), our proposal for fake ID detection achieves 13.91\% and 0\% EERs at the patch and the whole ID level, showing a good generalization to other databases. In addition to the path-based methodology introduced and the new FakeIDet method based on it, another key contribution of our article is the release of the first publicly available database that contains 48,400 patches from real and fake IDs, called FakeIDet-db, together with the experimental framework\footnote{\href{https://github.com/BiDAlabFakeIDet}{https://github.com/BiDAlab/FakeIDet-db}}. 
\end{abstract}    
\section{Introduction}
\label{sec:intro}

Verification of the veracity of digital content is one of the great challenges facing society today \cite{Aslett2024}. With the rapid advances made in the field of Generative AI \cite{surv_gen_diff_models}, it is easy to synthesize non-existent content \cite{Melzi_2023_ICCV, shahreza2025hyperface}, or to modify existing one \cite{PERNUS2025111022}. Although these methods can be used for good purposes, e.g., correct biases \cite{melzi2023synthetic, pena25bias} or improve performance in some scenarios with limited data \cite{DEANDRESTAME2025103099,2023_IJCB_SynFacePAD2023_Fang, sfdr_challenge}, they can also be used for malicious purposes such as generating injurious DeepFakes \cite{rathgeb2022handbook, TOLOSANA2020131} or misinformation \cite{robust_domain_misinformation}. In particular, the synthesis of non-existent fake IDs or the manipulation of real ones has started gaining attention due to their impressive realism \cite{mudgalgundurao2022pixel, tapia2024first}. This is strongly related to the area of \textit{presentation attacks} in biometrics \cite{2013ICBfaceSpoofComp,2023_IJCB_SynFacePAD2023_Fang,2023_Book-PAD_Face_JHO}.

Recent news has revealed how fake identity documents (IDs) are used nowadays for several frauds, such as underage purchases of alcohol\footnote{\href{https://www.nytimes.com/2025/02/13/nyregion/students-high-tech-fake-ids.html}{https://www.nytimes.com/2025/02/13/nyregion/students-high-tech-fake-ids.html}} or impersonating to open accounts in digital services such as crypto-exchanges or digital banks\footnote{\href{https://www.404media.co/inside-the-underground-site-where-ai-neural-networks-churns-out-fake-ids-onlyfake/}{https://www.404media.co/inside-the-underground-site-where-ai-neural-networks-churns-out-fake-ids-onlyfake/}}, which use \gls{kyc} verification systems. This problem has been exacerbated by the rapid advance of Artificial Intelligence (AI) and Generative AI, which have allowed one to buy fake ID documents (identity card, passport, driving license, etc.) on websites for a reasonable price (20\$), making this a critical challenge in terms of security. Few studies have preliminarily analyzed the problem of fake ID detection, proposing very valuable ideas and resources \cite{synth_id_card_db, gonzalez2025forged,mudgalgundurao2022pixel}. However, there are several limitations in the field that need to be covered to properly advance this research line.

On the one hand, there are no public databases that contain real (a.k.a. \textit{bona fide}) data. This is mainly produced due to privacy concerns, as the information included in IDs is very sensitive. Previous studies in the literature have always considered private in-house databases not publicly available \cite{synth_id_card_db, gonzalez2025forged,mudgalgundurao2022pixel}. This results in two main limitations: \textit{i)} the lack of standard benchmarks to properly compare novel approaches with the state of the art in fake ID detection, and \textit{ii)} the poor performance of existing fake ID detectors due to the lack of official real data for training machine learning-based approaches, as can be seen in a recent international challenge carried out at the IJCB 2024 conference \cite{tapia2024first}.



\begin{figure*}[!ht]
	\centering
	\includegraphics[width=\linewidth]{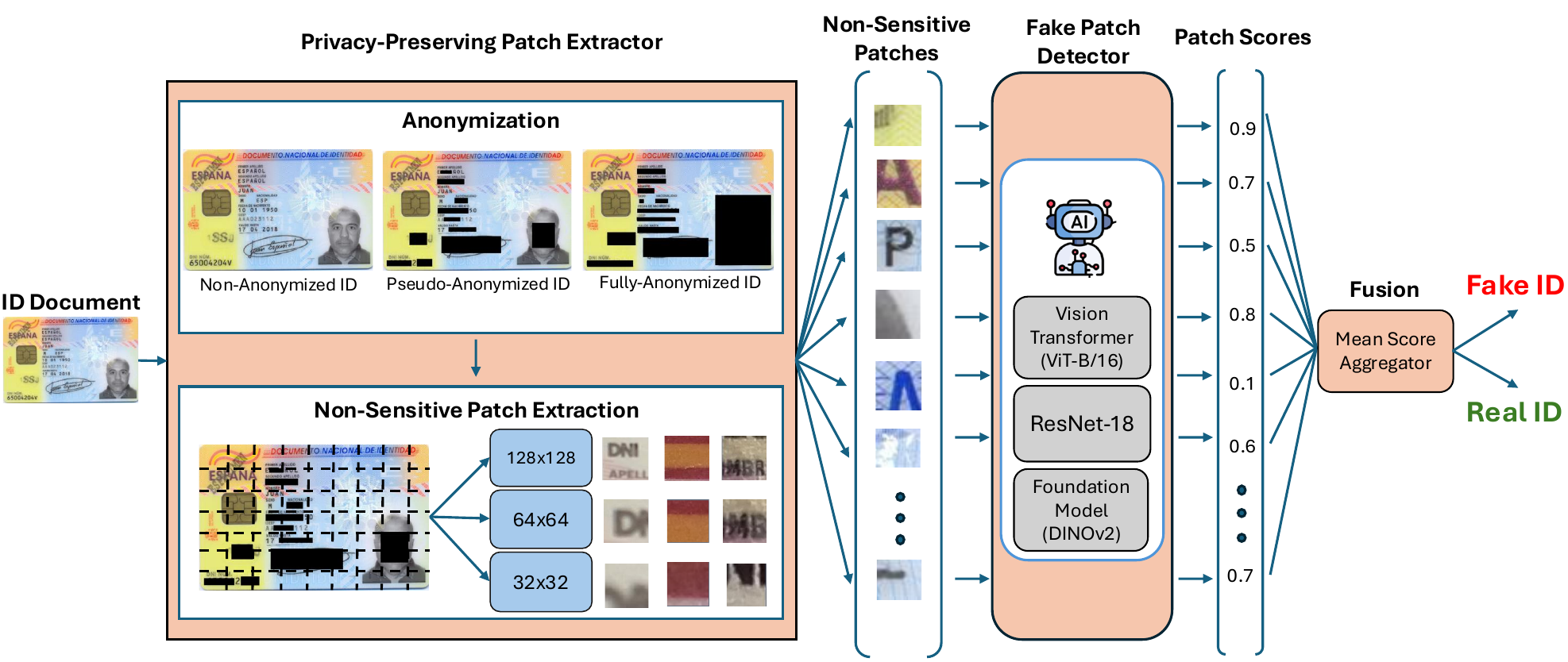}
	\caption{Graphical representation of FakeIDet, our proposed patch-wise approach for privacy-preserving fake ID detection, exploring a trade-off between performance and privacy (i.e., amount of sensitive data available as input to the model).}
	\label{fig:graph-abstract}
\end{figure*}

One of the first databases introduced in the literature that includes different types of fake IDs is the MIDV database family \cite{arlazarov2019midv, bulatov2020midv, Bulatov2022}. With the original purpose of Optical Character Recognition (OCR), the authors synthetically created a set of physical IDs and passports. They used several digital templates from multiple countries that they filled with names and addresses from Wikipedia and artificially generated faces. The authors increased the number of fake IDs with time as they released different versions of the databases. Later, Polevoy \textit{et al.} used in \cite{polevoy2022document} the documents of the MIDV-family to create fake documents considering three different \gls{pai} species: \textit{color print}, \textit{gray print}, and \textit{screen} \cite{2023_Book-PAD_Face_JHO}. They also provided real samples, although they should be considered fake documents, as they are not real, just prints in higher quality (i.e., \textit{glossy print}). Recently, Benalcazar \textit{et al.} proposed in \cite{synth_id_card_db} the use of \gls{gan} to create synthetic Chilean IDs. The proposed GAN was trained using only real Chilean ID samples. Although the data synthesized by the authors are not strictly real ID samples (note that GAN traces can be used to spot these synthetic samples \cite{2020_JSTSP_GANprintR_Neves}), this was presented as a good idea to partially cover the lack of real data, for example, as a data augmentation strategy. In \cite{gonzalez2025forged}, Gonzalez and Tapia proposed a more sophisticated fake ID generation, printing on a \gls{pvc} card, which resembles even more the appearance of real IDs.

All the variability mentioned in terms of digital/physical attacks results in very poor fake ID detection performances, especially in unconstrained scenarios. One of the first fake ID detection approaches was presented in \cite{mudgalgundurao2022pixel}, where the authors trained a \gls{cnn} to perform pixel-wise classification to detect different types of \gls{pai}s. The authors created their own private in-house database with both real and fake documents, including print and screen PAIs. Gonzalez and Tapia proposed in \cite{gonzalez2025forged} a two-stage system which first uses a neural network to evaluate if the submitted ID card is real or fake, concretely using digital PAIs of type \textit{composite} or \textit{synthetic}. The second network classified the ID considered between real or fake for physical PAIs such as \textit{printing, screen} or \textit{\gls{pvc}}. 

Finally, we would like to highlight the recent fake ID contest celebrated in IJCB 2024 \cite{tapia2024first}. Due to privacy concerns, organizers did not provide participants with a database to train the proposed fake ID detectors, stressing the difficulty of the challenge. The evaluation of the detectors submitted was done by the organizers using a private sequestered database. Regarding performance, very poor results were achieved by all teams in the competition, where the winner of the competition achieved an \gls{eer} of 21.87\%. This result highlights how challenging fake ID detection is in realistic settings.

\begin{figure*}[t]
    \centering
    \begin{subfigure}{0.25\linewidth}
        \centering
        \includegraphics[width=\textwidth]{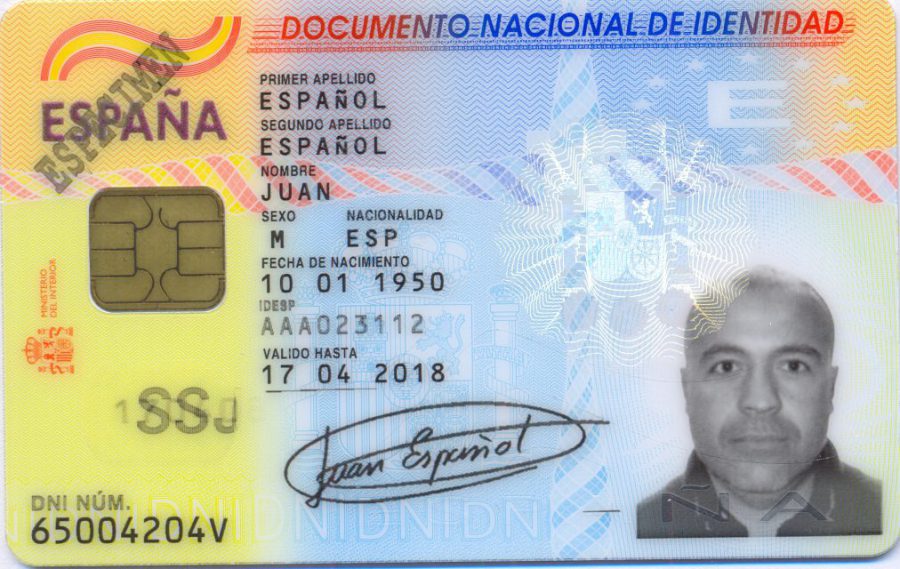}
        \caption{Non-anonymized ID.}
        \label{fig:semi-anon}
    \end{subfigure}
    \quad
    \quad
    \quad
    \begin{subfigure}{0.25\linewidth}
        \centering
        \includegraphics[width=\textwidth]{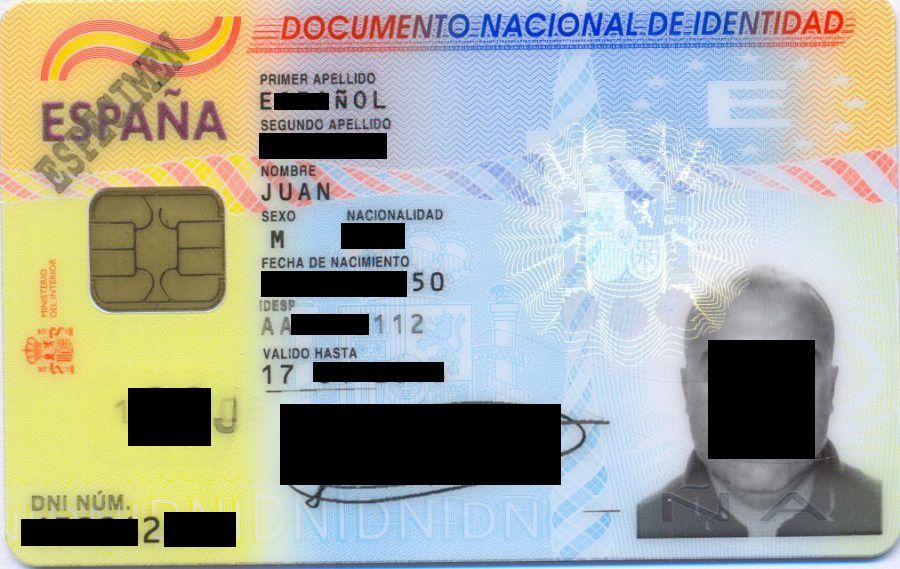}
        \caption{Pseudo-Anonymized ID.}
        \label{fig:pseudo-anon}
    \end{subfigure}
    \quad
    \quad
    \quad
    \begin{subfigure}{0.25\linewidth}
        \centering
        \includegraphics[width=\textwidth]{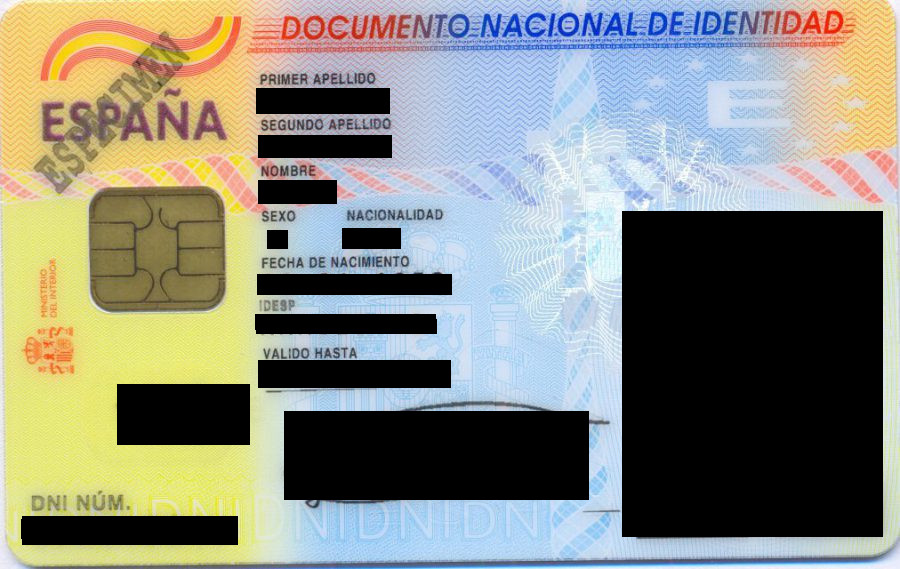}
        \caption{Fully-Anonymized ID.}
        \label{fig:fully-anon}
    \end{subfigure}
    \caption{Graphical examples of the three different anonymization configurations considered in our proposed method, with different levels of privacy in terms of the sensitivity of the information available in the ID.}
        \label{fig:anon_non-anon}
\end{figure*}

In this paper, we propose to tackle the task of fake ID detection from a different perspective, exploring a novel trade-off between performance and privacy. Concretely, instead of considering the whole ID as input to the models, our proposed approach explores how privacy can be enhanced through: \textit{i)} two levels of anonymization for an ID (i.e., fully- and pseudo-anonymized), and \textit{ii)} different patch size configurations, varying the amount of sensitive data visible in the patch image. These patches are then introduced into a deep learning model, classifying them as real or fake. Fig. \ref{fig:graph-abstract} provides a representation of our proposed approach. Our main contributions are as follows:

\begin{itemize}
    \item We propose a novel patch-wise approach for privacy-preserving fake ID detection, called FakeIDet, exploring several configurations in terms of performance and privacy. In particular, as can be seen in Fig. \ref{fig:graph-abstract}, we consider three different scenarios in terms of anonymization (non-, pseudo-, and fully-anonymized ID) and also in terms of the size of the patches ($128\times128$, $64\times64$, and $32\times32$), providing different levels of privacy and amount of training samples depending on the particular scenario.
    \item We provide the first publicly available database that contains patches from both real and fake IDs, called FakeIDet-db. The proposed database contains 30 real IDs from 30 different subjects. In addition, for each real ID, different methods have been used in order to produce high-quality fake IDs, considering different PAI species (i.e., print and screen). The pseudo- and fully-anonymized ID data with the best patch size configuration ($64\times64$) will be available on our GitHub\footnote{\href{https://github.com/BiDAlabFakeIDet}{https://github.com/BiDAlab/FakeIDet-db}} for research purposes.
    \item In addition, we explore different state-of-the-art deep learning methods such as Residual Network (ResNet) \cite{he2016deep}, Vision Transformer (ViT) \cite{dosovitskiy2020image}, or DINOv2 Foundation Model \cite{oquab2023dinov2} for the extraction of discriminative real/fake patches, considering pre-trained models and fine-tuning. 

\end{itemize}


The remainder of the paper is organized as follows. In Sec. \ref{sec:proposed_method} we explain the details of our proposed patch-wise approach for privacy-preserving fake ID detection, FakeIDet. Sec. \ref{sec:proposed_db} provides the acquisition details of our new fake ID database, FakeIDet-db. In Sec. \ref{sec:experiments}, we explain the experimental framework of the study, including the experimental protocol and evaluation metrics. Sec. \ref{sec:prel_analysis} shows the results achieved by our proposed method. Finally, the conclusions are presented in Sec. \ref{sec:conclusion}.

\section{FakeIDet: Proposed Method}
\label{sec:proposed_method}

Fig. \ref{fig:graph-abstract} provides a graphical representation of FakeIDet, our proposed patch-wise approach for privacy-preserving fake ID detection. As can be seen, it comprises two main modules: \textit{i)} a privacy-preserving patch extractor, and \textit{ii)} a fake patch detector. Finally, a score fusion of the individual patches is considered to detect a whole ID as real or fake. We provide next all the details.

\subsection{Privacy-Preserving Patch Extractor}
\label{sub:patch_extractor}
Given an ID, we first explore three different scenarios in terms of anonymization, considering different configurations in terms of privacy, as can be seen in Fig. \ref{fig:anon_non-anon}:

\begin{enumerate}
    \item \textbf{Non-Anonymized ID}: there is no anonymization. All information included in the ID is available to detect real/fake IDs.
	\item \textbf{Pseudo-Anonymized ID}: sensitive fields are partially anonymized, such as the expiration date, ID number, the face of the owner and the support number, leaving some sections of information available. The name and surname are never displayed together (i.e., one of them is always anonymized). In addition, they are partially occluded, showing only some random characters, so that it is impossible to know the full name/surname of the owner.
	\item \textbf{Fully-Anonymized ID}: all fields with sensitive information about the owner (i.e., text, face image, and handwritten signature) are completely anonymized and only patches from outside of the sensitive information are collected.
\end{enumerate}

\begin{figure}[t]
    \centering
    \hspace{-0.5cm}
    \includegraphics[width=0.8\linewidth]{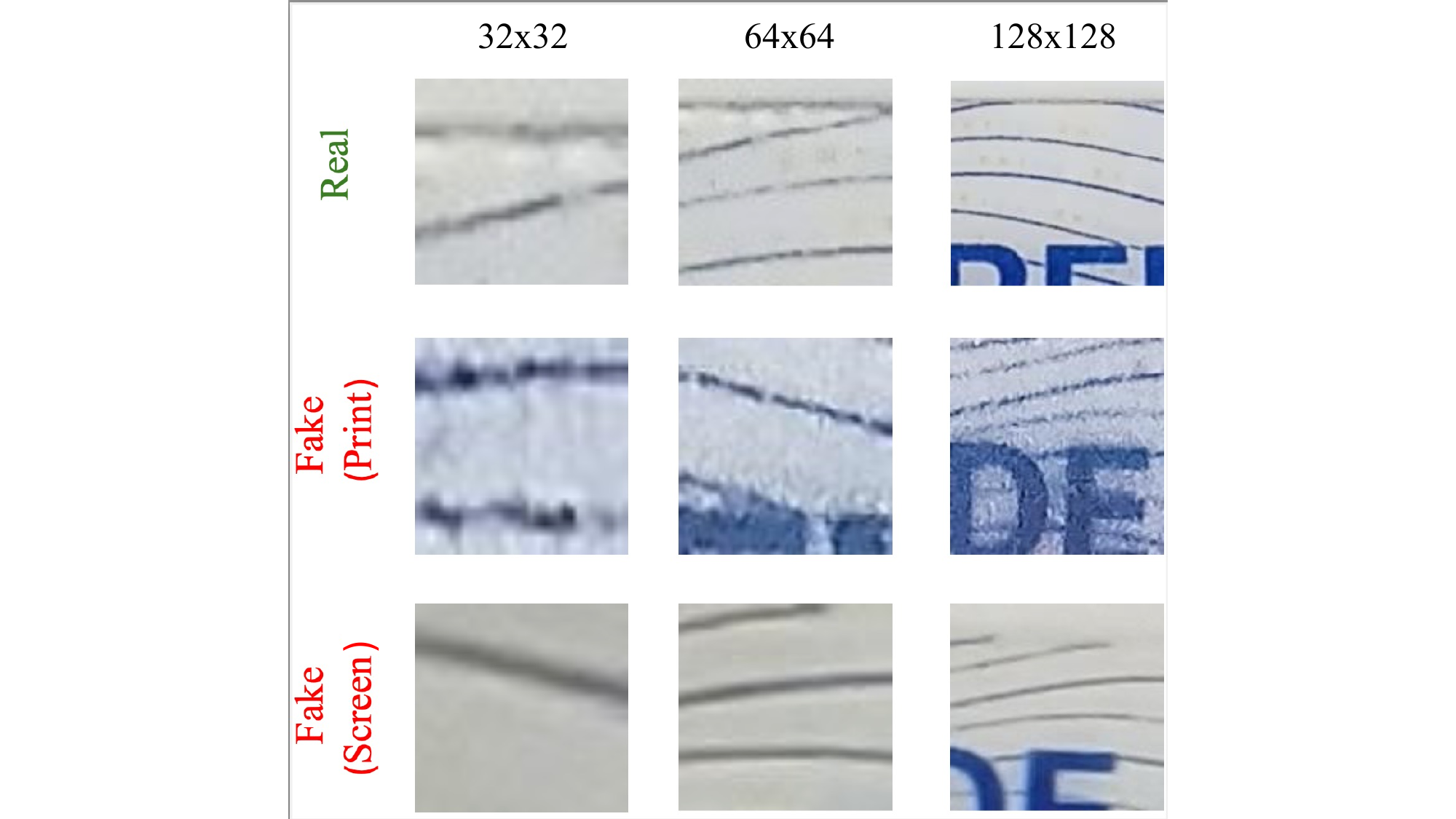}
    \caption{Graphical representation of real (top, green color) and fake (middle and bottom, red color) patches at the different sizes considered in the analysis.}
    \label{fig:patches_types}
\end{figure}

In addition to the anonymization scenario, we explore the extraction of patches of different sizes: $128\times128$, $64\times64$, and $32\times32$ pixels. The main motivation for this is also to increase privacy, showing less sensitive information of the subject as the image size of the patch is reduced. Graphical examples of real and fake patches of different sizes can be seen in Fig. \ref{fig:patches_types}. 

\subsection{Fake Patch Detector}
\label{sub:patch_detector}
After the extraction of patches from an ID, we train three different state-of-the-art deep learning models to classify each patch between real and fake.

\begin{itemize}
    \item \textbf{ResNet-18} \cite{he2016deep}: the ResNet architecture is based on convolutional layers, introducing for the first time residual connections to overcome the vanishing gradient problem in very deep models. The ResNet-based model considered in this study has been pre-trained with the popular ImageNet database \cite{ImageNet}.
    \item \textbf{ViT-B/16 Vision Transformer} \cite{dosovitskiy2020image}: the ViT was the first architecture to use the self-attention layers introduced in Transformers \cite{vaswani2017attention} for image classification. Concretely, we consider the ViT-based model pre-trained in the ImageNet database \cite{ImageNet}.
    \item \textbf{DINOv2} \cite{oquab2023dinov2}: this model uses the ViT architecture to learn visual feature representations, considering a self-supervised approach based on distillation \cite{hinton2015distilling}. This is based on a teacher-student scheme with a slight modification. While the student is fed with the whole image and local patches of it, the teacher model received only the whole image. The core idea of this training procedure is that the teacher model tries to match the embedding distribution of the student. This patch strategy, combined with centering and momentum updates for teacher model weights, helped DINO learn semantic representations without labels. 
\end{itemize}

The selected deep learning models have been modified in order to adapt them to our problem, fake ID detection. First, for both ResNet-18 and ViT-B/16 models pre-trained with ImageNet database, we replaced the final softmax layer (1,000 classes) with a fully connected layer based on a sigmoid activation in order to provide a continuous score for real (values close to 0) and fake (values close to 1) IDs. Regarding DINOv2, as it was trained using an unsupervised approach with the purpose of extracting visual features, we only added a fully connected layer with a sigmoid activation. Regarding the fine-tuning, we decided to keep the backbones of the models frozen and only train the new fully connected layers, given the limited amount of training data. For any patch size, the input was resized to $224\times224$, which is the input shape of the three models. The loss function selected to train these models was \gls{bce}. The selected optimizer was Adam \cite{2015-kingma} with a learning rate of $\alpha = 0.00015$ and exponential decays for the momentum estimators of $\beta_1 = 0.9$ and $\beta_2 = 0.999$. All models were trained for 150 epochs, but with an early stopping condition that ended the training if no improvements were obtained in the loss value for the validation split after 12 epochs. 

Finally, although our proposed method is designed to detect each patch as real or fake, it also allows for whole ID classification. This is referred to throughout the paper as ``Patch Level" or ``ID Level". Concretely, for the ``ID Level", in the present study, we consider a score fusion of the individual patches based on the mean of the predictions, obtaining a final score between 0 and 1. Other approaches such as majority voting could be used for the fusion, although this type of mechanism would only provide a binary output (not continuous value) with less discriminative information of the decision.
\section{FakeIDet-db: Proposed Database}
\label{sec:proposed_db}

\subsection{Real and Fake IDs: Acquisition}
\label{sub:id_acdquisition}
Due to the lack of publicly available databases, one of the key contributions of our study is the acquisition and publication of a new database, FakeIDet-db, comprising real and fake identification documents. For real IDs, there are a total of 30 images of Spanish IDs, each one belonging to a different subject (that is, 30 subjects in total). Spanish electronic IDs have changed over the years, which was also taken into account when capturing the data, considering three different versions/templates in the database. Referring to the capture device, we used a Redmi 9C NFC smartphone, a low-end device with a sensor of 13 MP, a wide $f$/2.2 aperture, which captures images in 4K resolution and 4:3 aspect with \gls{hdr}. All pictures were saved in JPEG format (very high quality, so compression artifacts are not noticeable, see, e.g., Fig.~\ref{fig:patches_types}). The photos were taken with a vertical distance of approximately 15 centimeters from the ID. 

Regarding fake IDs, we consider two types of PAIs: \textit{print} and \textit{screen}. In both cases, in order to generate high-quality fake IDs, we first scanned the real IDs using an HP ScanJet 8270 scanner at 600 \gls{dpi}. After getting the corresponding digital copies for each ID, we create the print \gls{pai} following a similar approach to \cite{arlazarov2019midv}. We used an EPSON ET-2850 Wi-Fi and regular paper to print the scanned versions of the real IDs, which then were laminated to improve the realism using a Fellowes Lunar A4 thermal laminator. Regarding the screen \gls{pai}, we display the ID images on a MacBook Pro 14" XDR display, with a resolution of $3024\times1964$, in full-screen mode and take the pictures so that the whole document displayed covered the whole camera preview of the smartphone. We selected this screen because it was the panel with more \gls{ppi} available (that is, 254 \gls{ppi}). This aspect is very important for high-quality fake IDs as the space between pixels in a screen is one of the most evident features of the screen \gls{pai} \cite{polevoy2022document}. In addition, we took special care to avoid any aliasing or interference patterns, such as Moiré patterns, when taking pictures. The proposed acquisition resulted in 90 real/fake IDs in total: 30 real IDs, 30 print fake IDs, and 30 screen fake IDs.

\subsection{Patch Generation}

As described in Sec. \ref{sec:proposed_method}, in this paper we explore the detection of fake IDs based on patches, containing different information and sizes depending on privacy restrictions. Regarding the pseudo- and fully-anonymized ID configurations, we covered the sensitive fields with pitch black rectangles (with the color code (0,0,0) in the RGB spectrum) using the GNU Image Manipulator Program (GIMP) and EasyOCR\footnote{\href{https://github.com/JaidedAI/EasyOCR}{https://github.com/JaidedAI/EasyOCR}}. After that, we used PyTorch’s unfold method to obtain the patches by specifying a stride with the same size as the patch. Patches with more than 90\% of its area with the (0,0,0) color in the RGB spectrum were discarded. Furthermore, the remaining patches were selected with a probability of $p=0.8$, making reconstruction even more difficult. Examples of real and fake patches of different sizes can be seen in Fig. \ref{fig:patches_types}, where we can see that the patches alone do not contain sensitive information.

In addition, as we plan to release the database, some mechanisms have been considered to avoid the reconstruction of the real IDs. First, only the pseudo- and fully-anonymized ID configurations are available for privacy reasons. In addition, patches from all real and fake IDs are randomized in terms of position and nomenclature. Finally, we remark that the extracted patches from a single ID are stored with a distinct code name, so that this approach enables both single patch and full ID evaluation.
\section{Experimental Framework}
\label{sec:experiments}

\subsection{Experimental Protocol}
\label{sub:exp_protocol}
The experimental protocol has been designed to analyze the feasibility of our proposed FakeIDet method.

\begin{table}
    \centering
    \setlength{\tabcolsep}{4pt} 
    \renewcommand{\arraystretch}{0.9} 
    \begin{tabular}{lrrrr}
        \toprule
        \textbf{Anon. Level} & \textbf{Full} & \textbf{128$\times$128} & \textbf{64$\times$64} & \textbf{32$\times$32} \\
        \midrule
        Non-Anon. & 60 & 9,520 & 39,440 & 144,160\\
        Pseudo-Anon. & 60 & 5,040 & 28,240 & 122,632 \\
        Fully-Anon. & 60 & 3,760 & 20,160 & 91,760 \\
        \bottomrule
    \end{tabular}
    \vspace{2mm}
    \caption{Number of data samples in our proposed database.}
    \label{tab:data_splits}
\end{table}

\begin{table*}[t]
   \centering
    \begin{tabular}{c|ccc|cccc}
        & \multicolumn{3}{c|}{\textbf{Patch Level}} & \multicolumn{4}{c}{\textbf{ID Level}} \\
        \toprule
        &  \textit{128$\times$128} & \textit{64$\times$64} & \textit{32$\times$32} & \textit{128$\times$128} & \textit{64$\times$64} & \textit{32$\times$32} & \textit{Full ID} \\
        \midrule
        \textit{ResNet-18} & 16.62 & 20.88 & 30.17 & \textbf{0.00} & \textbf{0.00} & 16.67 & 45.83 \\
        \textit{ViT-B/16} & 18.61 & 19.31 & \textbf{21.51} & \textbf{0.00} & \textbf{0.00} & \textbf{0.00} & 33.3 \\
        \textit{DINOv2} & \textbf{13.81} & \textbf{14.98} & 24.14 & \textbf{0.00} & \textbf{0.00} & \textbf{0.00} & 33.3 \\
        \midrule
        \shortstack{\textit{Avg.} \textit{\gls{eer}}} & \textbf{16.34} & 18.39 & 25.27 & \textbf{0.00} & \textbf{0.00} & 5.55 & 37.48\\
        
    \end{tabular}
    \vspace{2mm}
    \caption{Performance in terms of EER (\%) of FakeIDet for the \textbf{different patch size configurations} and deep learning models. Evaluations at both patch and ID levels are considered. For completeness, we also include in the ``Full ID" column the results achieved for the case of introducing the whole ID picture to the deep learning models, instead of patches, as this is the most popular method in the literature.}
    \label{tab:exp1_results_patch}
\end{table*}

First, in Sec. \ref{sub:patch_size_vs_det_perf} and Sec. \ref{sub:anal_sensitive}, experiments are carried out considering our novel database. The purpose of this analysis is to evaluate our proposed method in terms of performance and privacy, comparing the results achieved with the traditional approach followed in the literature, i.e., introducing the whole ID to the deep learning models. 

Regarding the data, we consider a balanced database composed of 30 real IDs and 30 fake IDs. Both types of PAIs (i.e., print and screen) are used to train our proposed fake ID detection method. Table \ref{tab:data_splits} provides a summary of the total number of IDs (shown as ``Full") and patches available depending on the anonymization configuration and the size of the patches. As can be seen, it is important to highlight that the amount of training data changes based on the selected privacy configuration, i.e., anonymization and patch size. Therefore, these experiments will evaluate the influence of anonymization and patch size configurations in the final trade-off of performance and privacy. Regarding the experimental protocol, our database is divided into the development (80\% of the real/fake IDs) and final evaluation (remaining 20\% of the real/fake IDs) dataset. This considers a realistic scenario with unseen IDs for testing (the first version of the Spanish ID is only seen in the final evaluation, not in training). Evaluation is performed at both patch and full ID levels.

After this first analysis using only our novel database, we evaluate in Sec. \ref{sec:cross_database_scenario} the performance of our proposed patch-wise approach for privacy-preserving fake ID detection under a cross-database scenario. This scenario aims to analyze the generalizability of FakeIDet to unseen databases, which is expected to be the typical scenario in real applications. In particular, the optimal configuration of our proposed method, discussed in Sec. \ref{sec:prel_anal_insights} and trained only on our new database is evaluated using PAIs from a different public database, DLC-2021 \cite{polevoy2022document}. This database includes physical PAIs such as print and screen, which are created following similar procedures as ours but captured under different conditions. For example, in DLC-2021, the acquisition is performed using different devices (i.e. Samsung S10 and iPhone XR vs. Redmi 9C NFC), distance, and light conditions, among others. More details and differences between both databases are included in Sec.~\ref{sec:cross_database_scenario}. Concretely, from DLC-2021 we consider 1,500 fake document samples (balanced among the different PAI species), which are processed considering the best patch configuration (see Sec. \ref{sec:prel_anal_insights}). The real patches are extracted from our novel database (i.e., the evaluation dataset) as there are no public databases that contain real ID samples. 

\subsection{Evaluation Metrics}
Similar to previous approaches presented in the literature \cite{tapia2024first}, we consider the ISO/IEC 30107-3 standard\footnote{\href{https://www.iso.org/standard/79520.html}{https://www.iso.org/standard/79520.html}} for the evaluation of fake ID detection technology: the \gls{bpcer} and the \gls{apcer}. The \gls{bpcer} metric tells how many bona-fide samples (i.e., real IDs) are incorrectly classified as attacks (i.e., fake IDs). The \gls{apcer} metric represents the same thing as \gls{bpcer} but for attacks (i.e. fake IDs) that are incorrectly classified as bona fide (i.e., real IDs). In addition, the Equal Error Rate (EER) metric, which gives the error rate at the operational point $\tau$ where the metrics $\textrm{\gls{bpcer}}(\tau)$ and $\textrm{\gls{apcer}}(\tau)$ are equal, is another popular metric considered in the literature to compare performance. Finally, as described in Sec. \ref{sec:proposed_method}, our proposed method is designed to detect real/fake patches as well as the whole ID. This is referred as ``Patch Level" or ``ID Level".

\section{Experimental Results}
\label{sec:prel_analysis}

We explore the feasibility of our proposed patch-wise approach for privacy-preserving fake ID detection. In particular, in Sec. \ref{sub:patch_size_vs_det_perf} we analyze the performance of FakeIDet in terms of patch size (i.e., $128\times128$, $64\times64$, and $32\times32$). Sec. \ref{sub:anal_sensitive} evaluates the performance of FakeIDet in terms of the anonymization setup (i.e., non-, pseudo-, and fully-anonymized ID). Then, in Sec. \ref{sec:prel_anal_insights}, we select the optimal configuration of FakeIDet in terms of performance and privacy. Finally, in Sec. \ref{sec:cross_database_scenario} we evaluate the performance of our optimal FakeIDet in a cross-database scenario. This scenario aims to analyze the generalizability of the proposed fake ID detection method to unseen PAIs and databases not considered for training. Evaluations at both the patch level and the identification document level are considered in the analysis.

\subsection{Patch Size vs. Detection Performance}
\label{sub:patch_size_vs_det_perf}

Table \ref{tab:exp1_results_patch} shows the performance in terms of EER (\%) of our FakeIDet method for the different patch size configurations and deep learning models. Evaluations are performed at both the patch level and the ID level. For completeness, we also include in the ``Full ID" column the results achieved for the case of introducing the whole ID picture to the models, instead of patches, as this is the most popular method in the literature. In this first analysis, we consider the case of having all information of the ID available (i.e., non-anomymized ID configuration).

\begin{table*}[t]
    \centering
    \begin{tabular}{c|ccc|ccc}
        & \multicolumn{3}{c|}{\textbf{Patch Level}} & \multicolumn{3}{c}{\textbf{ID Level}} \\
        \midrule
        & \textit{Non-Anon} & \textit{Pseudo-Anon} & \textit{Fully-Anon} & \textit{Non-Anon} & \textit{Pseudo-Anon} & \textit{Fully-Anon} \\
        \midrule
        \textit{ResNet-18 ($128\times128$)} & 16.62 & \textbf{16.80} & \textbf{11.83} & 0.00 & 0.00 & 0.00 \\
        \textit{ViT-B/16 ($128\times128$)} & 18.61 & 19.72 & 20.97 & 0.00 & 0.00 & 0.00 \\
        \textit{DINOv2 ($128\times128$)} & \textbf{13.81} & 17.76 & 15.59 & 0.00 & 0.00 & 0.00 \\
        \midrule
        \textit{Avg. EER  ($128\times128$)} & 16.34 & 18.09 & 16.13 &  0.00 & 0.00 & 0.00 \\
        \midrule
        \midrule
        \textit{ResNet-18 ($64\times64$)} & 20.88 & 22.44 & 22.60 & 0.00 & 0.00 & 0.00 \\
        \textit{ViT-B/16 ($64\times64$)} & 19.31 & 20.08 & 21.51 & 0.00 & 0.00 & 0.00 \\
        \textit{DINOv2 ($64\times64$)} & \textbf{14.98} & \textbf{15.38} & \textbf{14.83} & 0.00 & 0.00 & 0.00 \\
        \midrule
        \textit{Avg. EER  ($64\times64$) } & 18.39 & 19.3 & 19.64 &  0.00 & 0.00 & 0.00 \\
        
    \end{tabular}
    \vspace{2mm}
    \caption{Performance in terms of EER (\%) of our FakeIDet method for the \textbf{different anonymization configurations} (i.e., non-, pseudo-, and fully-anonymized ID), the two best patch size configurations ($128\times128$ and $64\times64$), and all deep learning models. Evaluations at both patch and ID level are considered.}
    \label{tab:exp1_sense_results}
\end{table*}

Analyzing the performance of our proposed method at the patch level, we can see that, in general, the best patch size configuration is $128\times128$, achieving an average EER of 16.34\%. The EER increases as the patch size is reduced, with values of 18.39\% and 25.27\% EER for the $64\times64$ and $32\times32$ configurations, respectively. These results are also interesting from the point of view of the amount of training data. For example, for the $128\times128$ configuration, in this experiment we have 9,520 patches in total, whereas for the $32\times32$ configuration this value increases to 144,160 patches. The results achieved in this particular experiment reject the hypothesis that more patches of smaller size may perform better than fewer patches of larger size. 

Also, it is important to highlight the different performance of deep learning models depending on the patch size configuration. DINOv2 achieves the best performance with 13.81\% EER for the $128\times128$ configuration, an absolute improvement of 2.79\% and 4.80\% EER compared to the ResNet-18 and ViT-B/16 models, respectively. This performance improvement may be produced due to the objective of the DINOv2 training procedure being to match the embedding distribution from images with full context to local patches. By following this procedure, DINOv2 is able to align both local and global features from a single image, and potentially benefits from local information given as patches in our proposed method.

It is also interesting to analyze the optimal trade-off between performance and privacy, as patch size directly contributes to privacy, since much or less information of the subject would be visible in patch images, as can be seen in Fig. \ref{fig:patches_types}. In this sense, it is surprising to see that DINOv2 performs quite similar for the $64\times64$ configuration, with an absolute increase in EER of just 1.17\% (13.81\% vs 14.98\%). A similar trend is observed for the ViT-B/16 model, where its difference in performance between patch size configurations is only 0.7\% EER (18.61\% vs. 19.31\%). These results suggest that deep learning models based on ViT architectures provide features that are more robust to patch size. In the case of ResNet-18, it achieves good performance for the $128\times128$ configuration, but lacks robustness when the patch size decreases, e.g., from 16.62\% to 30.17\% EER for the $128\times128$ and $32\times32$ configurations, respectively.

Finally, in Table \ref{tab:exp1_results_patch} we include the performance of the \mbox{FakeIDet} method at the whole ID level. For completeness, we also include in the ``Full ID" column the results achieved for the case of introducing the whole ID picture to the deep learning models, instead of patches, as this is the most popular method in the literature. In general, very good results can be achieved using our proposed method, with 0\% EER results in most configurations and deep learning models. These results are much better compared to the ``Full ID" traditional case considered in the state of the art, e.g., for the case of DINOv2 the EER increases to 33.3\% for the ``Full ID" scenario. These results confirm the feasibility of our proposed patch-wise approach for privacy-preserving fake ID detection.

\subsection{Anonymization vs. Detection Performance}
\label{sub:anal_sensitive}

In addition to the different patch size configurations studied before, the present section analyzes the effect of the different anonymization configurations in the final performance. Similarly to the previous analysis, the aim is to obtain a trade-off between performance and privacy. 

Table \ref{tab:exp1_sense_results} shows the performance in terms of EER (\%) of our FakeIDet method for the different anonymization configurations (i.e., non-, pseudo-, and fully-anonymized ID), and the two best patch size configurations ($128\times128$ and $64\times64$), and all deep learning models. Evaluations are considered at both the patch and the ID levels.

Analyzing the different anonymization configurations, in general we can observe in all deep learning models an EER increasing from the non-anonymized to the fully-anonymized configurations, e.g., for the $64\times64$ patch size, the EER increases on average from 18.39\% to 19.64\% EER. These results make sense as more patches and with additional fake patterns in the sensitive information are available for the non-anonymized configuration. Nevertheless, the performance is in general similar for all anonymization configurations, being possible to reduce the amount of sensitive information to detect if an ID is real or fake, e.g., for the DINOv2 model and $64\times64$ patch size, EER values of 14.98\% and 14.83\% are achieved for the non- and fully-anonymized configurations, respectively. 


Finally, for completeness, we also include in Table \ref{tab:exp1_sense_results} the performance of our proposed method at the whole ID level. Again, very good results can be achieved using our proposed method in all anonymization configurations, with 0\% EER results in both $128\times128$ and $64\times64$ patch size configurations and all deep learning models. These results confirm again the feasibility of our proposed patch-wise approach for privacy-preserving fake ID detection.

\subsection{Proposed Method: Optimal Configuration}
\label{sec:prel_anal_insights}
The results achieved in Sec. \ref{sub:patch_size_vs_det_perf} and Sec. \ref{sub:anal_sensitive} have proved the feasibility of our FakeIDet method compared to the traditional one, that is, introducing the whole ID picture into the deep learning model. Several configurations have been studied in terms of patch size and anonymization with the purpose of selecting an optimal trade-off between performance and privacy due to the sensitive information included in IDs.

Taking this into account, we have decided to select as our optimal configuration the DINOv2 model with the patch size $64\times64$ and the fully-anonymized ID setup. This is a privacy-preserving configuration, as sensitive information is not considered in the analysis. In addition, as can be seen in Fig. \ref{fig:patches_types}, images with a patch size of $64\times64$ reveal much less information on the subject than the $128\times128$ configuration. For this particular configuration, DINOv2 achieves EER values of 14.83\% at the patch level and 0\% at the ID level.

\begin{figure}[t]
	\centering
	\begin{subfigure}{0.23\linewidth}
		\includegraphics[width=\linewidth]{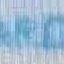}
		\label{fig:dlc_patch_1}
	\end{subfigure}
	\hfill
	\begin{subfigure}{0.23\linewidth}
		\includegraphics[width=\linewidth]{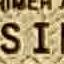}
		\label{fig:dlc_patch_2}
	\end{subfigure}
        \hfill
        \begin{subfigure}{0.23\linewidth}
		\includegraphics[width=\linewidth]{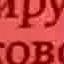}
		\label{fig:dlc_patch_3}
	\end{subfigure}
	\hfill
	\begin{subfigure}{0.23\linewidth}
		\includegraphics[width=\linewidth]{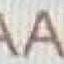}
		\label{fig:dlc_patch_4}
	\end{subfigure}
	\caption{Examples of fake patches from DLC-2021 \cite{polevoy2022document}.}
    \label{fig:dlc_patches_examples}
\end{figure}

\subsection{Cross-Database Scenario}
\label{sec:cross_database_scenario}

This section evaluates the generalizability of our proposed FakeIDet method to unseen PAIs and databases not considered for training, which is expected to be the typical scenario in real applications. Concretely, we consider the optimal configuration described in Sec. \ref{sec:prel_anal_insights}, trained only with our novel database. For the final evaluation, we consider fake IDs from a different database, DLC-2021 \cite{polevoy2022document}. This database includes physical PAIs such as print (\textit{glossy}, \textit{color}, and \textit{gray}) and \textit{screen}. Graphical examples of $64\times64$ fake patches extracted from DLC-2021 are included in Fig. \ref{fig:dlc_patches_examples}. Table \ref{tab:dlc_results} shows the performance in terms of \gls{eer} (\%) of DINOv2 ($64\times64$, fully-anonymized ID) for this database.

Before analyzing the results, we would like to remark the difficulty of the scenario as: \textit{i)} different types of physical PAIs are considered in the evaluation of the fake ID detection method, \textit {ii)} different types of templates and documents are considered in the analysis (ID card and passport), and from different countries such as Albania, Findland, Estonia, etc., and \textit{iii)} different types of acquisitions are considered in terms of smartphones (iPhone XR y Samsung S10), resolution, distance of the camera, angles, etc. Additionally, we would like to remark that the DLC-2021 database contains Spain IDs, but their templates belong to the first version, which is not seen while training our proposed method, as commented in the experimental protocol, see Sec. \ref{sub:exp_protocol}.

\begin{table}[t]
    \centering
    \begin{tabular}{c|c|c}
        \textbf{PAI Class} & \textbf{Patch Level} & \textbf{ID Level} \\
        \midrule
        \textit{Glossy-print} & 13.33 & 0.00\\
        \textit{Color-print} & 12.02 & 0.00\\
        \textit{Gray-print} & 12.99 & 0.00\\
        \textit{Screen} & 17.29 & 0.00\\
        \midrule
        \textit{Avg. EER (\%)} & \textbf{13.91}\% & \textbf{0.00}\%
    \end{tabular}
    \vspace{2mm}
    \caption{Performance in terms of \gls{eer} (\%) of our FakeIDet method ($64\times64$, fully-anonymized ID) \textbf{for the DLC-2021 \cite{polevoy2022document} database, which is not considered for training our proposed method}. Fake IDs from different PAIs are considered in the analysis.}
    \label{tab:dlc_results}
\end{table}

As can be seen in Table \ref{tab:dlc_results}, our proposed FakeIDet method is able to generalize well to patches extracted from different PAIs. In particular, the proposed method achieves on average 13.91\% EER for the analysis at patch level, similar to the performance achieved in our database (i.e., 14.83\% EER). This suggests that when physical PAIs are used, our method remains robust, effectively detecting fake patches from different distributions. Finally, if we analyze the performance at the whole ID level, we can see that our proposed method is able to detect real/fake IDs without mistakes (0\% EER), proving a good generalization ability.

\section{Conclusion}

\label{sec:conclusion}
We presented FakeIDet, a novel patch-wise approach for privacy-preserving fake ID detection and explored several configurations in terms of performance and privacy. Due to the lack of public databases that contain both real and fake IDs, we acquired a new database (FakeIDet-db), which includes real IDs from 30 subjects plus fake IDs using print and screen methods. 

Through an in-depth experimental framework, we have validated our proposed method considering intra- and cross-database scenarios. In particular, our optimal configuration is based on the DINOv2 model, with a configuration based on $64\times64$ patches and fully-anonymized ID. With this setup and over a different database not seen in training (DLC-2021), our proposed method has been able to achieve EER values of 13.91\% and 0\% for the analysis at patch level and the whole ID level, respectively, proving to generalize well to other PAIs and conditions. However, we are aware that these good results might be produced due to the lack of public databases, as they do not cover all possible real-life scenarios. Future work will be oriented towards increasing the size and variability of our public database.
\section*{Acknowledgements}
\label{sec:acks}

Support by INTER-ACTION (PID2021-126521OBI00 MICINN/FEDER), M2RAI (PID2024-160053OB-I00 MICIU/FEDER), Cátedra ENIA UAM-VERIDAS en IA Responsable (NextGenerationEU PRTR TSI-100927-2023-2), FakeIDet (FUAM-Veridas R\&D Agreement signed July 2025), and PowerAI+ (SI4/PJI/2024-00062 Comunidad de Madrid and UAM).

{
	\small
	\bibliographystyle{ieee}
	\bibliography{main}
}

\end{document}